\documentclass[letterpaper, 10 pt, conference]{ieeeconf}  

\IEEEoverridecommandlockouts                              

\title{\LARGE \bf
Deformable Gaussian Splatting for Efficient and High-Fidelity Reconstruction of Surgical Scenes
}

\author{Jiwei Shan$^{1*}$, Zeyu Cai$^{2*}$, Cheng-Tai Hsieh$^{3}$, Shing Shin Cheng$^{1}$ and Hesheng Wang$^{3}$
\thanks{*The first two authors contributed equally. This work was supported in part by the Natural Science Foundation of China under Grant 62361166632, 62225309, 62073222, and U21A20480, in part by Innovation and Technology Commission of Hong Kong (ITS/235/22) and in part by Multi-scale Medical Robotics Center, InnoHK. Corresponding Authors: Hesheng Wang, Shing Shin Cheng.}
\thanks{$^{1}$ Department of Mechanical and Automation Engineering and T Stone Robotics Institute, The Chinese University of Hong Kong, Hong Kong.}%
\thanks{$^{2}$ Department of Electronic Engineering, Shanghai Jiao Tong University, Shanghai
200240, China.}%
\thanks{$^{3}$ Department of Automation, Shanghai Jiao Tong University, Shanghai
200240, China}
}

\usepackage{amssymb}
\usepackage{graphicx}
\usepackage{cite}
\usepackage{multirow}
\usepackage{booktabs}
\usepackage{amsmath}
\usepackage{graphicx}
\usepackage{subfigure}
\usepackage{textcomp}
\usepackage{color}
\usepackage{tabularx}
\usepackage{multirow}
\usepackage{xcolor}
\definecolor{gray}{rgb}{0.5, 0.5, 0.5} 
\usepackage[colorlinks, linkcolor=blue, anchorcolor=blue, citecolor=blue]{hyperref}
\begin{document}

\maketitle
\thispagestyle{empty}
\pagestyle{empty}

\begin{abstract}
Efficient and high-fidelity reconstruction of deformable surgical scenes is a critical yet challenging task. Building on recent advancements in 3D Gaussian splatting, current methods have seen significant improvements in both reconstruction quality and rendering speed. However, two major limitations remain: (1) difficulty in handling irreversible dynamic changes, such as tissue shearing, which are common in surgical scenes; and (2) the lack of hierarchical modeling for surgical scene deformation, which reduces rendering speed. To address these challenges, we introduce EH-SurGS, an efficient and high-fidelity reconstruction algorithm for deformable surgical scenes. We propose a deformation modeling approach that incorporates the life cycle of 3D Gaussians, effectively capturing both regular and irreversible deformations, thus enhancing reconstruction quality. Additionally, we present an adaptive motion hierarchy strategy that distinguishes between static and deformable regions within the surgical scene. This strategy reduces the number of 3D Gaussians passing through the deformation field, thereby improving rendering speed. Extensive experiments demonstrate that our method surpasses existing state-of-the-art approaches in both reconstruction quality and rendering speed. Ablation studies further validate the effectiveness and necessity of our proposed components. We will open-source our code upon acceptance of the paper.
\end{abstract}

\section{INTRODUCTION}
Reconstructing deformable surgical scenes from stereoscopic endoscopic videos is a critical task. High-quality reconstruction enhances surgeons' understanding of anatomical structures, thereby improving the success rate and safety of procedures. Additionally, in medical education, clear surface details and accurate reconstruction of soft tissue structures contribute to creating a high-quality virtual surgical environment. This provides doctors with a realistic and safe training platform, helping them master essential skills. However, achieving high-quality surgical scene reconstruction faces significant challenges. Due to physiological movements of the body (such as breathing) and interactions between surgical instruments and soft tissues (such as pulling and shearing), the topological structure of the surgical scene typically changes over time. Therefore, it is crucial to develop an effective method for deformable scene representation.
\begin{figure}[t]
\centerline{\includegraphics[width=1\linewidth]{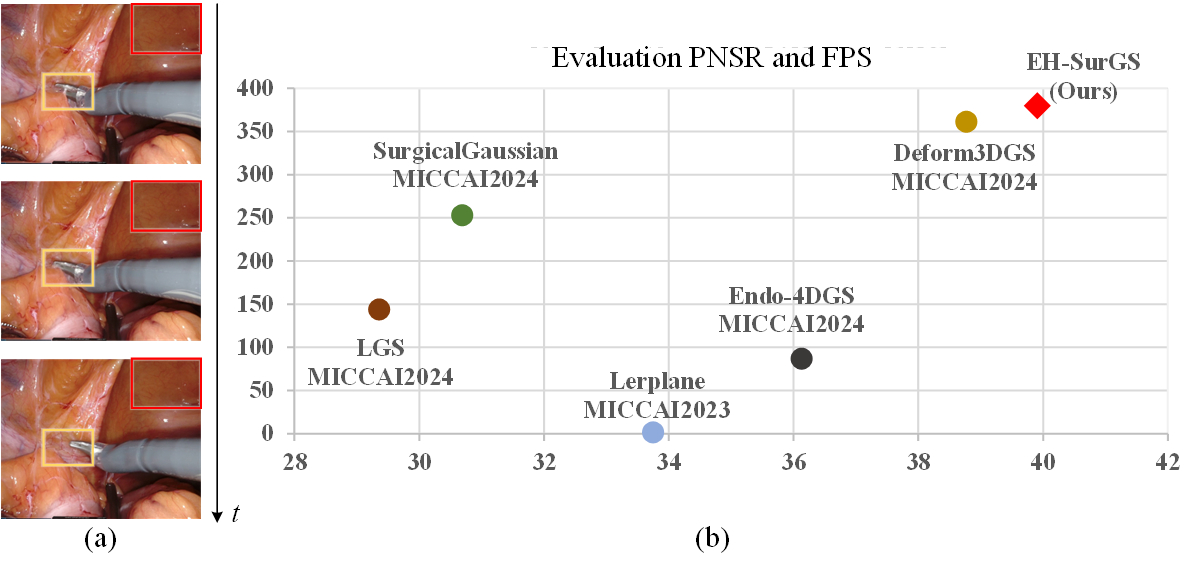}}
    \vspace{-12pt}
    \caption{(a) Visualization of tissue shearing (yellow box) and static areas (red box) in a surgical scene. (b) Comparison of reconstruction quality (PSNR) and rendering speed (FPS) with state-of-the-art surgical scene reconstruction algorithms. Our method achieves state-of-the-art performance.}
    \vspace{-20pt}
    \label{fig1}
\end{figure}

In recent years, neural rendering has advanced significantly \cite{tewari2020state,tewari2022advances,mildenhall2021nerf}. Several methods \cite{endonerf,endosurf,yang2023neural,batlle2023lightneus,10542414} have used implicit neural representations to reconstruct surgical scenes, achieving impressive results. However, these methods require dense sampling of millions of rays to render images. They cannot be rendered in real-time and struggle to meet the demands of real medical applications. Recently, some methods \cite{yang2024deform3dgs,huang2024endo,li2024endosparse,liu2024lgs,xie2024surgicalgaussian} have adopted 3D Gaussian splatting (3DGS) \cite{kerbl20233d} to overcome these limitations. These methods use 3D Gaussians for canonical space representation, combined with a deformation field to model deformable surgical scenes. 
These approaches have significantly improved training time, rendering speed, and reconstruction quality compared to implicit neural representations. However, they still face challenges.

First, in surgical scenes, soft tissues are often sheared from interactions between surgical instruments and tissues, as shown in Fig.~\ref{fig1}(a) (yellow boxes). These dynamic changes are usually irreversible. They cannot be effectively modeled by existing deformation methods because there is no clear correspondence between the states before and after deformation. Secondly, existing methods struggle to distinguish between different scales of motion in various regions of the scene. Specifically, as shown in Fig.~\ref{fig1}(a) (red boxes), some areas in the surgical scene remain static over time. To obtain a scene at a specific timestamp, existing methods typically warp all 3D Gaussians in the canonical space through the deformation field, even for those that do not change. This process slows down rendering speed.

To address the above problems, we propose EH-SurGS, an \underline{\textbf{e}}fficient and \underline{\textbf{h}}igh-fidelity reconstruction algorithm for deformable surgical scenes. First, to model the irreversible dynamic changes caused by intraoperative operations (such as shearing), we introduce the concept of the life cycle for 3D Gaussians. The life cycle activates each 3D Gaussian during specific periods for rendering. During other periods, it is disabled. By introducing this concept, EH-SurGS effectively models both general and irreversible deformations in surgical scenes, improving reconstruction performance. Secondly, we propose an adaptive motion hierarchy strategy that distinguishes between static and deformable areas in surgical scenes. This significantly reduces the number of 3D Gaussians that need warping, improving rendering speed. Finally, quantitative and qualitative results on multiple endoscopy datasets show that EH-SurGS outperforms state-of-the-art methods in reconstruction quality and rendering speed. Ablation studies further validate the effectiveness and necessity of our proposed components. In summary, our contributions are as follows:
\begin{enumerate}
    \item To model the irreversible dynamic changes in surgical scenes, the concept of life cycle for 3D Gaussians is introduced. This offers a more comprehensive representation of deformable scenes and demonstrates high-fidelity reconstruction performance.
    \item To improve rendering speed, an adaptive motion hierarchy strategy is proposed. It effectively distinguishes static areas from deformable areas in surgical scenes.
    \item Extensive experiments show that our method outperforms existing methods in reconstruction quality and rendering speed. Ablation studies also validate the effectiveness and necessity of the proposed components.
\end{enumerate}
\section{Related Works}
\subsection{Deformable Surgical Scene Reconstruction Based on Neural Implicit Representation}
Neural implicit representation is a novel method for scene representation \cite{mildenhall2021nerf,tewari2022advances}. It uses multi-layer perceptrons (MLPs) to capture scene geometry and appearance. Recent work has extended this approach to dynamic scenes, including deformable surgical scenes \cite{endonerf,endosurf,batlle2023lightneus,yang2023neural}. EndoNeRF\cite{endonerf} is the first to apply neural implicit representation to deformable surgical scene modeling. It uses two MLPs to represent the typical radiance field and time-varying deformation network, achieving remarkable results. EndoSurf \cite{endosurf} builds on EndoNeRF, modeling scene geometry using signed distance fields. Forplane \cite{yang2024efficient} improves training efficiency by classifying the four-dimensional space into multiple orthogonal two-dimensional feature planes. LightNeus \cite{batlle2023lightneus} focuses on scene lighting characteristics and designs a static scene surface reconstruction algorithm. Despite these advancements, neural implicit representation methods require intensive acquisition and querying of millions of rays through the network. This process consumes significant computational resources, reduces rendering speed, and is unsuitable for actual medical applications.
\subsection{Deformable Surgical Scene Reconstruction Based on 3D Gaussian Splatting} 
3D Gaussian splatting (3DGS) \cite{kerbl20233d} introduces an efficient differentiable rendering scheme based on tile rasterization, resulting in significantly faster rendering. Due to its superior performance, many works use 3D Gaussians representations for the canonical space, combined with a deformation field, to model deformable surgical scenes. The deformation field is modeled in different ways. Specifically, EndoSparse \cite{li2024endosparse} and SurgicalGaussian \cite{xie2024surgicalgaussian} use MLPs to model scene deformation, similar to EndoNeRF \cite{endonerf}. LGS \cite{liu2024lgs} and Endo-4DGS \cite{huang2024endo} model soft tissue deformation using multiple orthogonal 2D feature planes and a small MLP. This method is similar to Forplane \cite{yang2024efficient} and can further reduce training time. Deform3DGS \cite{yang2024deform3dgs} represents deformable surgical scenes using explicit basis functions. These methods outperform previous neural implicit representation methods in training time, rendering speed, and reconstruction quality. However, these methods do not account for soft tissue shearing during surgery or the phenomenon of motion hierarchy in the scene. In this paper, by introducing the concept of the life cycle of 3D Gaussians and combining it with an adaptive motion hierarchy strategy, we further improve the reconstruction performance and rendering speed.
\section{Method}
\begin{figure*}[t]
    \centerline{\includegraphics[width=1\linewidth]{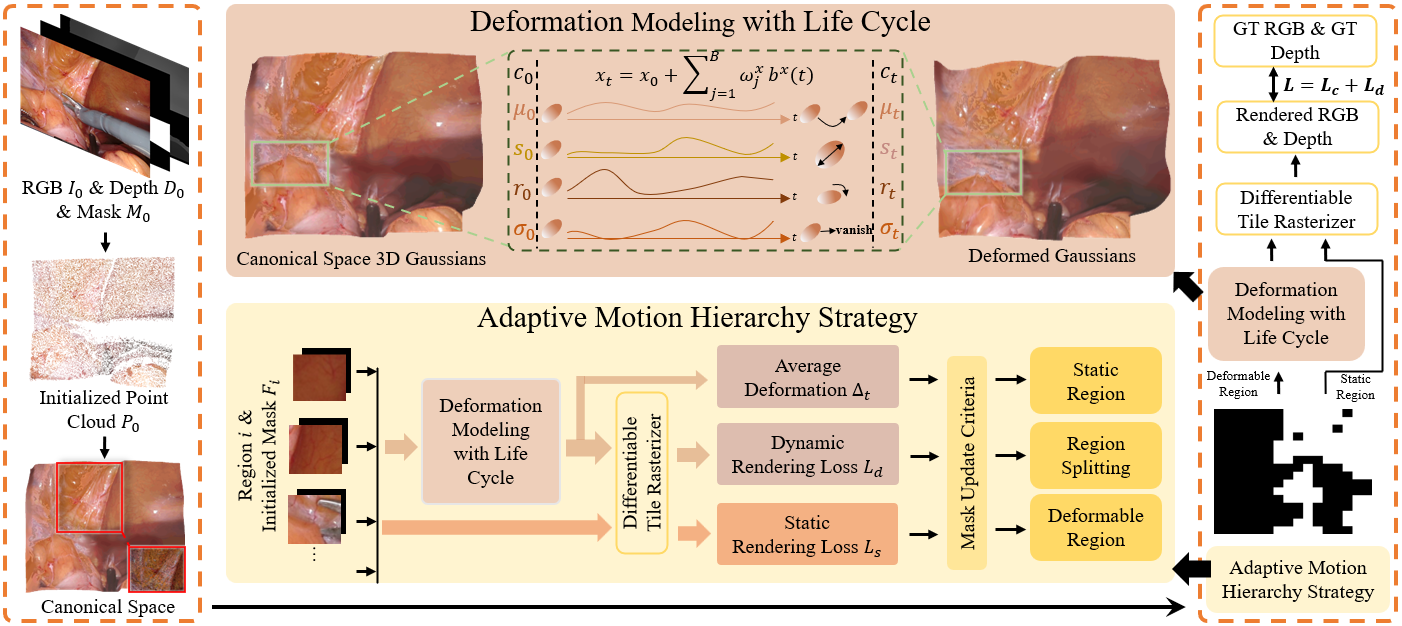}}
    \vspace{-8pt}
    \caption{Overview of EH-SurGS. It consists of two core modules: Deformation Modeling with Life Cycle (Sec. \ref{sec:def}) and Adaptive Motion Hierarchy Strategy (Sec. \ref{sec:mask}). EH-SurGS initializes the point cloud \(P_0\) through back-projection based on the input RGB image, depth map, and surgical tool mask. \(P_0\) is used to initialize 3D Gaussians to represent the canonical space. The Adaptive Motion Hierarchy Strategy is then applied to distinguish between deformable and static regions. For deformable regions, Deformation Modeling with Life Cycle is used to obtain the deformed Gaussians. Finally, RGB and depth are rendered through the differentiable tile rasterizer, and the loss is computed by comparing the rendered results with the inputs.}
    \vspace{-15pt}
    \label{fig:overview}
\end{figure*}

Given a set of videos of surgical scenes captured by a stereo camera, our goal is to develop a model based on 3D Gaussian splatting \cite{kerbl20233d} with high-quality scene reconstruction performance and fast rendering capability for deformable surgical scenes. The pipeline of EH-SurGS is shown in Fig.~\ref{fig:overview}. EH-SurGS takes a sequence of frame data $\{I_i, D_i, M_i, P, t_i\}_{i=1}^T$ as input. Here, $I_i \in \mathbb{R}^{H \times W \times 3}$ and $D_i \in \mathbb{R}^{H \times W}$ represent the left RGB image and depth map of the $i$-th frame, respectively. $M_i \in \mathbb{R}^{H \times W}$ is the surgical tool mask, which excludes unwanted pixels belonging to the surgical tool. $P \in \mathbb{R}^{4 \times 4}$ is the projection matrix. $t_i = i/T$ is the normalized timestamp for each frame.

EH-SurGS consists of three main components: 3D Gaussian canonical space, deformation modeling with life cycle, and an adaptive motion hierarchy strategy. The key techniques of 3DGS are reviewed in detail in Sec.~\ref{sec:3dgs}. We use 3D Gaussians to represent the canonical space, taking advantage of their fast rendering speed. Sec.~\ref{sec:def} describes the deformation modeling module and explains how the life cycle concept is introduced to model irreversible deformation. Sec.~\ref{sec:mask} explains how to distinguish static areas from deformable ones in the surgical scene using an adaptive motion hierarchy strategy. Finally, Sec.~\ref{sec:opt} presents the overall workflow and optimization process of our algorithm.
\subsection{Preliminary: 3D Gaussian Splatting}\label{sec:3dgs}
3D Gaussian Splatting \cite{kerbl20233d} is an explicit representation of 3D scenes that uses anisotropic 3D Gaussian functions to model static 3D scenes. Each Gaussian function is defined by the following formula:
\begin{equation}
G(x)=e^{-\frac{1}{2} (x-\mu)^T \Sigma^{-1} (x-\mu)}.\label{eq1}
\end{equation}
where $\Sigma \in \mathbb{R}^{3 \times 3}$ is an anisotropic covariance matrix, and $\mu \in \mathbb{R}^3$ is its mean vector. To ensure a physically meaningful covariance matrix during optimization, $\Sigma$ is parameterized using a scale matrix $S$ and a rotation matrix $R$:
\begin{equation}
\Sigma = R S S^T R^T.\label{eq2}
\end{equation}
Here, $S$ is stored by its diagonal elements $s = \operatorname{diag}\left(s_x, s_y, s_z\right)$, and $R$ is constructed from a unit quaternion $r \in \mathbb{R}^4$. Additionally, 3D Gaussians include opacity $\sigma \in \mathbb{R}$ and spherical harmonics (SH) coefficients to represent view-dependent colors $c \in \mathbb{R}^3$.

During rendering, 3D Gaussians are projected into 2D image space using splatting techniques \cite{zwicker2002ewa}. With the viewing transformation $W$ and the Jacobian of the affine approximation of the projective transformation $J$, the 2D covariance matrix in camera coordinates is computed as:
\begin{equation}
\Sigma^{\prime}=J W \Sigma W^T J^T.\label{eq3}
\end{equation}

The color $\hat{C}(p)$ and depth $\hat{D}(p)$ of pixel $p$ in the rendered image are determined through an alpha blending process. This process blends $N$ visible Gaussians sorted by depth:
\begin{equation}
\hat{C} = \sum_{i \in N} c_i \alpha_i \prod_{j=1}^{i-1}\left(1 - \alpha_j\right), \quad
\hat{D} = \sum_{i \in N} d_i \alpha_i \prod_{j=1}^{i-1}\left(1 - \alpha_j\right)
\label{equ:rgb&depth} 
\end{equation}
\begin{equation}
    \alpha_i = \sigma_{i} e^{-\frac{1}{2}(p - \mu_i)^T {\Sigma^{\prime}}^{-1} (p - \mu_i)},
    \label{equ:alpha}
\end{equation}
where $d_i$ represents the $i$-th Gaussian's z-depth coordinate in view space. Finally, under the supervision of rendering loss, the attributes of each Gaussian are optimized. The 3D scene is thus represented by the parameter set of 3D Gaussians, i.e., $P = \{G: \mu, s, r, \sigma, c\}$. During optimization, 3D Gaussian splatting also performs adaptive density control \cite{kerbl20233d}, further enhancing geometric shapes and rendering quality.
\subsection{Deformation Modeling with Life Cycle}\label{sec:def}
Following existing methods, we use the 3DGS introduced in Sec.~\ref{sec:3dgs} to model the canonical space, taking full advantage of its superior performance. For the deformation field, whether modeled through MLPs \cite{li2024endosparse,liu2024lgs,xie2024surgicalgaussian}, 2D feature planes \cite{huang2024endo}, or explicit motion basis \cite{yang2024deform3dgs}, only general deformation can be modeled. For irreversible dynamic changes, such as shearing in surgical scenes, the above methods usually produce artifacts. To address this problem, we propose life cycle-based deformation modeling.

Specifically, we use learnable Gaussian functions and weights to represent the transformation of position \(\mu\), rotation \(r\), and scale \(s\) of 3D Gaussians, simulating deformation in the scene. The Gaussian function is defined as:
\begin{equation}
b(t) = \exp\left( -\frac{1}{2\sigma^2}(t - \theta)^2 \right) 
\label{equ:gau}
\end{equation}
where \(\theta\) and \(\sigma\) represent the center and variance of the Gaussian basis function, which are learnable parameters. The position \(\mu_t\), rotation \(r_t\), and scale \(s_t\) of the 3D Gaussians at time \(t\) are determined by:
\begin{equation}
    x_t = x_0 + \sum_{j=1}^{B} \omega_j^{x} b^{x}(t)
    \label{equ:x}
\end{equation}
where \(x_t\) refers to \(\mu_t\), \(r_t\), and \(s_t\). \(B\) is the number of selected Gaussian basis functions, and in this paper, \(B=20\). \(\omega_j^{x}\) are the learnable weights. The position, rotation, and scale at each time \(t\) are expressed by the parameter sets: \(\Theta_{\mu} = \{\omega_{\mu}, \theta_{\mu}, \sigma_{\mu}\}\), \(\Theta_{r} = \{\omega_{r}, \theta_{r}, \sigma_{r}\}\), and \(\Theta_{s} = \{\omega_{s}, \theta_{s}, \sigma_{s}\}\).

To model the life cycle of each 3D Gaussian, STG \cite{li2024spacetime} and GaussianPrediction \cite{zhao2024gaussianprediction} use the product of the opacity \(\sigma\) of the 3D Gaussians in canonical space and the temporal radial basis function \cite{li2024spacetime} or sigmoid function \cite{zhao2024gaussianprediction}. However, experiments show that this approach introduces artifacts in depth rendering (see Sec.~\ref{sec:ablation}). In this paper, we apply the same method as for modeling position changes:
\begin{equation}
    \alpha_t = \alpha_0 + \sum_{j=1}^{B} \omega_j^{\alpha} b^{\alpha}(t)
    \label{equ:alpha}
\end{equation}

Modeling the life cycle controls whether the 3D Gaussians in the canonical space are rendered during specific time periods. Our experiments show that this simple yet effective method models the disappearance of content in the surgical scene, reducing 3D Gaussian residuals and improving rendering quality (see Sec.~\ref{sec:ablation}).
\subsection{Adaptive Motion Hierarchy Strategy}\label{sec:mask}
Although 3DGS-based deformable surgical scene modeling methods improve rendering speed compared to previous approaches, they generally use a simple deformation rendering strategy. These methods fail to differentiate between motion scales in different regions of the scene, leading to reduced rendering efficiency. To solve this, we propose an adaptive motion hierarchy strategy. Specifically, we use a mask \(F \in \mathbb{R}^{H \times W}\)with the same resolution as the image to distinguish deformable from static areas. The mask \(F\) is updated alternately with the optimization of EH-SurGS. We explain this in two parts: update criteria and update process.

\textbf{Update criteria.} Deformable and static regions are distinguished from two perspectives. The first is average deformation. After the 3D Gaussians in the canonical space pass through the deformation modeling module, we calculate the position changes \(\Delta x\), \(\Delta y\), and \(\Delta z\) for all 3D Gaussians in each region. These changes are normalized, summed, and averaged to obtain the average deformation \(\Delta_t\) for each region. If \(\Delta_t\) exceeds a set threshold \(\delta_1\) (experimentally determined as 0.05), the region is added to the potential dynamic region set \(Q\). Otherwise, it is placed in the potential static region set \(W\). The second perspective is dynamic and static rendering loss. For the same region, we calculate the rendering loss of the 3D Gaussians with and without the deformation module, denoted as \(L_d\) and \(L_s\). If the region is static, \(L_d\) and \(L_s\) should be consistent {(difference less than the threshold \(\delta_2\), experimentally determined as 0.5),} and the region is added to the potential static region set \(W'\). Conversely, if the region is dynamic, \(L_d\) should be much smaller than \(L_s\), and the region is added to the dynamic region set \(Q'\).

\textbf{Update process.} The following steps outline the process for updating mask assignments to regions: 
\begin{enumerate}
    \item \textbf{Initialization.} Divide the input RGB image into \(N \times N\) regions. The mask \(F_i\) for each region \(i\) is initially set as dynamic, i.e., \(F_i = 0\). 
    \item \textbf{Update the mask.} When the number of iterations reaches the preset threshold \(N_m\), calculate the average deformation \(\delta_t\), \(L_d\), and \(L_s\) for each region. Update the mask of each region based on the update criteria. The intersection of the potential static region sets \(W\) and \(W'\) defines a static region, while the intersection of the potential dynamic region sets \(Q\) and \(Q'\) defines a dynamic region. 
    \item \textbf{Region splitting.} To improve accuracy, the initially divided regions are split based on specific criteria. For a region \(q\), if it belongs to conflicting sets, i.e., \(q \in W\) and \(q \in Q'\), or \(q \in W'\) and \(q \in Q\), the region is split. The splitting operation divides the original region evenly into four dynamic blocks, ensuring more detailed separation. 
    \item \textbf{Update \(N_m\).} In the early stages of training, frequent mask updates can destabilize training and negatively affect reconstruction quality (Sec.~\ref{sec:ablation}). Therefore, we dynamically update \(N_m\) based on the model's optimization progress:
    \begin{equation}
        N_m = N_m' \times \text{factor}, \quad \text{factor} = \frac{L_l}{L_c}
    \end{equation}
    where \(N_m\) and \(N_m'\) represent the updated and previous thresholds, respectively. \(L_c\) and \(L_l\) denote the current and previous losses after the mask update, as calculated by \eqref{equ:loss}.
\end{enumerate}

The updated mask will control whether the 3D Gaussians in the canonical space need to be deformed in the subsequent training process.
\begin{table*}[t]
\caption{Mean quantitative evaluation across different datasets (variance in parentheses). The symbol $\uparrow$ indicates that higher values correspond to better performance, and $\downarrow$ indicates the opposite. The bold font highlights the best results.}\label{tab:qua}
    \renewcommand{\arraystretch}{0.6} 
    \setlength{\tabcolsep}{3pt} 
    \vspace{-4pt}
\begin{tabularx}{\textwidth}{c*{12}{>{\centering\arraybackslash}X}}
    \toprule
    \multirow{2}{*}{Methods} & 
    \multicolumn{5}{c}{EndoNeRF-Cutting} & \multicolumn{5}{c}{EndoNeRF-Pulling} \\
    \cmidrule(lr){2-6} \cmidrule(lr){7-11}
    & PSNR\(\uparrow\) & SSIM\(\uparrow\) & LPIPS\(\downarrow\) &FPS\(\uparrow\) & Time(sec)\(\downarrow\) & PSNR\(\uparrow\) & SSIM\(\uparrow\) & LPIPS\(\downarrow\) & FPS \(\uparrow\) & Time(sec)\(\downarrow\)\\
    \midrule
    Forplane\cite{yang2023neural}&33.75(.011)&0.900(.000)&0.114(.001)&1.75(0.02)&179(3.7)&36.28(.009)&0.936(.000)&0.085(.000)&1.44(0.02)&186(2.0)  \\
 Deform3DGS\cite{yang2024deform3dgs}&38.77(.008)&0.967(.000)&0.042(.000)&361.33(2.05)&\textbf{91(4.7)}
 &38.33(.012)&0.961(.000)&0.063(.000)&361.33(4.78)&\textbf{90(0.5)}\\
    Endo-4DGS\cite{huang2024endo}&36.13(.241)&0.951(.000)&0.054(.000)&87.00(4.67)&250(4.9)
                                  &37.18(.342)&0.955(.004)&0.072(.007)&97.67(1.25)&243(10.0)  \\
    LGS\cite{liu2024lgs}&29.36(.017)&0.937(.003)&0.089(.000)&144.02(0.57)&120(1.0)&31.90(.009)&0.940(.000)&0.100(.001)&105.16(10.58)&113(10.6)\\
    SurgicalGaussian \cite{xie2024surgicalgaussian}&30.69(.029)&0.896(.000)&0.202(.001)&253.00(13.44)&160(1.6)&31.98(.065)&0.912(.000)&0.219(.002)&200.67(14.70)&154(0.3) \\
    EH-SurGS (ours)&\textbf{39.91(.014)}&\textbf{0.972(.000)}&\textbf{0.034(.000)}&\textbf{379.67(4.02)}&105(0.9)&\textbf{38.72(.042)}&\textbf{0.963(.000)}&\textbf{0.062(.001)}&\textbf{387.00(5.35)}&97(1.5)  \\
    \midrule
    \multirow{2}{*}{Methods} & 
    \multicolumn{5}{c}{Hamlyn-Clip1} & \multicolumn{5}{c}{Hamlyn-Clip2} \\
    \cmidrule(lr){2-6} \cmidrule(lr){7-11}
    & PSNR\(\uparrow\) & SSIM\(\uparrow\) & LPIPS\(\downarrow\) & FPS\(\uparrow\) & Time(sec)\(\downarrow\)& PSNR\(\uparrow\) & SSIM\(\uparrow\) & LPIPS\(\downarrow\) & FPS \(\uparrow\) & Time(sec)\(\downarrow\)\\
    \midrule
    Forplane\cite{yang2023neural}&35.02(.460)&0.934(.002)&0.060(.002)&11.27(0.50)&173(0.8)&38.49(1.26)&0.958(.003)&\textbf{0.037(.004)}&17.15(0.23)&155(1.3)\\
    Deform3DGS \cite{yang2024deform3dgs}&37.31(.059)&0.973(.000)&0.068(.000)&357.67(2.49)&\textbf{100(1.7)}&40.76(.168)&0.983(.000)&0.061(.001)&374.33(2.87)&\textbf{89(2.1)}  \\
    Endo-4DGS \cite{huang2024endo}&34.87(.215)&0.963(.000)&0.087(.000)&94.00(3.56)&250(5.8)&40.97(.050)&0.984(.000)&{0.043(.001)}&87.00(2.94)&250(3.6)\\
    LGS \cite{liu2024lgs}&26.94(.008)&0.917(.000)&0.172(.000)&148.79(2.82)&142(1.3)&30.55(.005)&0.944(.000)&0.143(.000)&150.44(0.77)&130(1.0)\\
    SurgicalGaussian\cite{xie2024surgicalgaussian}&30.41(.162)&0.942(.001)&0.182(.004)&376.33(10.62)&155(3.5)&36.43(.211)&0.976(.000)&0.090(.001)&390.33(15.15)&150(2.5)\\
    EH-SurGS (ours) &\textbf{38.80(.082)}&\textbf{0.976(.000)}&\textbf{0.054(.001)}&\textbf{379.33(1.89)}&115(2.1)&\textbf{42.02(.159)}&\textbf{0.984(.000)}&0.052(.000)&\textbf{400.00(5.89)}&107(2.2)\\
        \midrule
    \multirow{2}{*}{Methods} & 
    \multicolumn{5}{c}{StereoMIS-P1\_1} & \multicolumn{5}{c}{StereoMIS-P1\_2} \\
    \cmidrule(lr){2-6} \cmidrule(lr){7-11}
    & PSNR\(\uparrow\) & SSIM\(\uparrow\) & LPIPS\(\downarrow\) & FPS\(\uparrow\) & Time(sec)\(\downarrow\)& PSNR\(\uparrow\) & SSIM\(\uparrow\) & LPIPS\(\downarrow\) & FPS \(\uparrow\) & Time(sec)\(\downarrow\)\\
    \midrule
    Forplane\cite{yang2023neural}&28.47(.018)&0.824(.000)&0.258(.000)&1.20(0.00)&215(1.8)&28.95(.068)&0.829(.000)&0.239(.001)&1.11(0.03)&229(1.0)\\
    Deform3DGS \cite{yang2024deform3dgs}&34.54(.012)&0.896(.000)&\textbf{0.170(.000)}&357.67(2.49)&\textbf{92(1.3)}&34.87(.000)&0.911(.000)&0.156(.000)&340.33(0.47)&\textbf{107(4.5)}\\
    Endo-4DGS \cite{huang2024endo}&33.52(.826)&0.884(.014)&0.180(.038)&95.67(0.47)&240(0.9)&34.18(.057)&0.904(.000)&\textbf{0.150(.003)}&98.33(0.47)&248(2.1)\\
    LGS \cite{liu2024lgs}&23.46(.000)&0.842(.000)&0.270(.000)&142.58(1.83)&144(0.4)&25.47(0.02)&0.819(.000)&0.315(.000)&146.43(3.80)&119(1.2)\\
    SurgicalGaussian \cite{xie2024surgicalgaussian}&28.91(.198)&0.831(.002)&0.366(.010)&304.00(12.36)&159(3.3)&31.26(.008)&0.859(.000)&0.251(.000)&237.67(10.96)&150(1.8)\\
    EH-SurGS (ours)&\textbf{34.73(.000)}&\textbf{0.898(.000)}&0.177(.001)&\textbf{380.33(2.49)}&111(2.7)&\textbf{35.09(.017)}&\textbf{0.913(.000)}&0.155(.000)&\textbf{349.00(1.41)}&131(1.1)\\
    \bottomrule
\end{tabularx}
\vspace{-15pt}
\end{table*}
\subsection{Learning the Model}\label{sec:opt}  
Our framework jointly optimizes the parameters \(P\) of 3D Gaussians in the canonical space and the parameters \(\Theta_{\mu}\), \(\Theta_{r}\), \(\Theta_{\alpha}\), and \(\Theta_{s}\) of the deformable model. Following \cite{yang2024deform3dgs}, before training begins, we generate a dense point cloud \(P_0\) from the input depth map and RGB image. This point cloud is used to initialize the attributes of the 3D Gaussians in the canonical space. If the iteration reaches the mask update threshold \(N_m\), the mask \(F\) is updated. Otherwise, the mask \(F\) determines which 3D Gaussians need to pass through the deformable model. These Gaussians obtain their position, rotation, scale, and opacity at time \(t\) through deformation modeling with the life cycle, using \eqref{equ:gau}, \eqref{equ:x}, and \eqref{equ:alpha}. The model then renders both RGB and depth maps using differentiable tile-based rasterization. Finally, the loss function is constructed by comparing the rendered results with the inputs, as follows:
\begin{equation}
    \mathcal{L}_C = \|(1-M) \odot (\hat{C} - C)\|, \quad \mathcal{L}_D = \|(1-M) \odot (\hat{D} - D)\|
    \label{equ:loss}
\end{equation}
where \(M\), \(C\), and \(D\) represent the input surgical tool mask, RGB image, and depth map. Following \cite{wang2023sparsenerf}, we incorporate a ranking loss \(\mathcal{L}_{\text{rank}}\). The full objective loss of EH-SurGS is formulated as \(L = L_C + L_D + \lambda \mathcal{L}_{\text{rank}}\), where \(\lambda = 0.0002\).

\section{Experiment}\label{sec:exp}
\textbf{Datasets.} We evaluate the performance of EH-SurGS using three endoscopic datasets with deformable tissues. The \textbf{EndoNeRF dataset} \cite{endonerf} is a robotic stereo video captured from a single viewpoint during a prostatectomy performed by the da Vinci robot. We use two publicly available videos. One video (Pulling) shows soft tissue being pulled by surgical tools, while the other (Cutting) involves soft tissue being cut. In addition to RGB images, the dataset includes surgical tool masks and dense depth maps. The \textbf{StereoMIS dataset} \cite{hayoz2023learning} is recorded using the da Vinci Xi surgical robot. We use two clips from videos P1\_1 and P1\_2, which contain natural deformations due to breathing and artificial deformations caused by surgical interactions. The \textbf{Hamlyn dataset} \cite{recasens2021endo}, provided by the Hamlyn Center at Imperial College London\footnote{http://hamlyn.doc.ic.ac.uk/vision/}, contains various challenging scenes. The rectified images, stereo depth, and camera calibration information are available from \cite{recasens2021endo}. We also use the widely adopted Segment Anything Model \cite{xiong2024efficientsam} to generate surgical instrument masks. Two clips from the video Rectified08 are used, referred to as Hamlyn-Clip1 and Hamlyn-Clip2 in the following sections. These clips show surgical instruments cutting soft tissue.

\textbf{Baselines and Metrics.}
We choose Forplane \cite{yang2024efficient}, one of the leading methods for deformable surgical scene reconstruction based on neural implicit representation.  In addition, we compare our method with state-of-the-art approaches based on 3D Gaussian splatting. These methods include Deform3DGS \cite{yang2024deform3dgs}, Endo-4DGS \cite{huang2024endo}, EndoSparse \cite{li2024endosparse}, LGS \cite{liu2024lgs}, and SurgicalGaussian \cite{xie2024surgicalgaussian}. For quantitative comparison, we use the peak signal-to-noise ratio (PSNR), structural similarity index (SSIM) \cite{ssim}, and learning-based perceptual image patch similarity (LPIPS) \cite{lpips} as metrics to evaluate reconstruction performance. Efficiency is assessed by measuring rendering speed (FPS, frames per second) and training time (seconds).

\textbf{Implementation Details.} We run EH-SurGS on a desktop PC equipped with an NVIDIA RTX 4090 GPU. Baselines and our method are trained for 3000 iterations using the Adam optimizer with an initial learning rate of $1.6 \times 10^{-3}$. To be fair, we run all the methods on a dataset three times and report the average results. During the mask $F$ update process, we set $N = 4$. Since all datasets are captured using fixed endoscopic cameras, the projection matrix $P$ is set as the identity matrix. The EndoNeRF-Cutting dataset contains 156 images, while the Pulling dataset contains 63 images. Hamlyn-Clip1 and Hamlyn-Clip2 contain 121 and 104 images, respectively. Stereo-P1-1 and P1-2 contain 184 and 70 images, respectively. We divide each dataset into training and testing sets with a 7:1 ratio.
\begin{figure*}[t]
    \centerline{\includegraphics[width=1\linewidth]{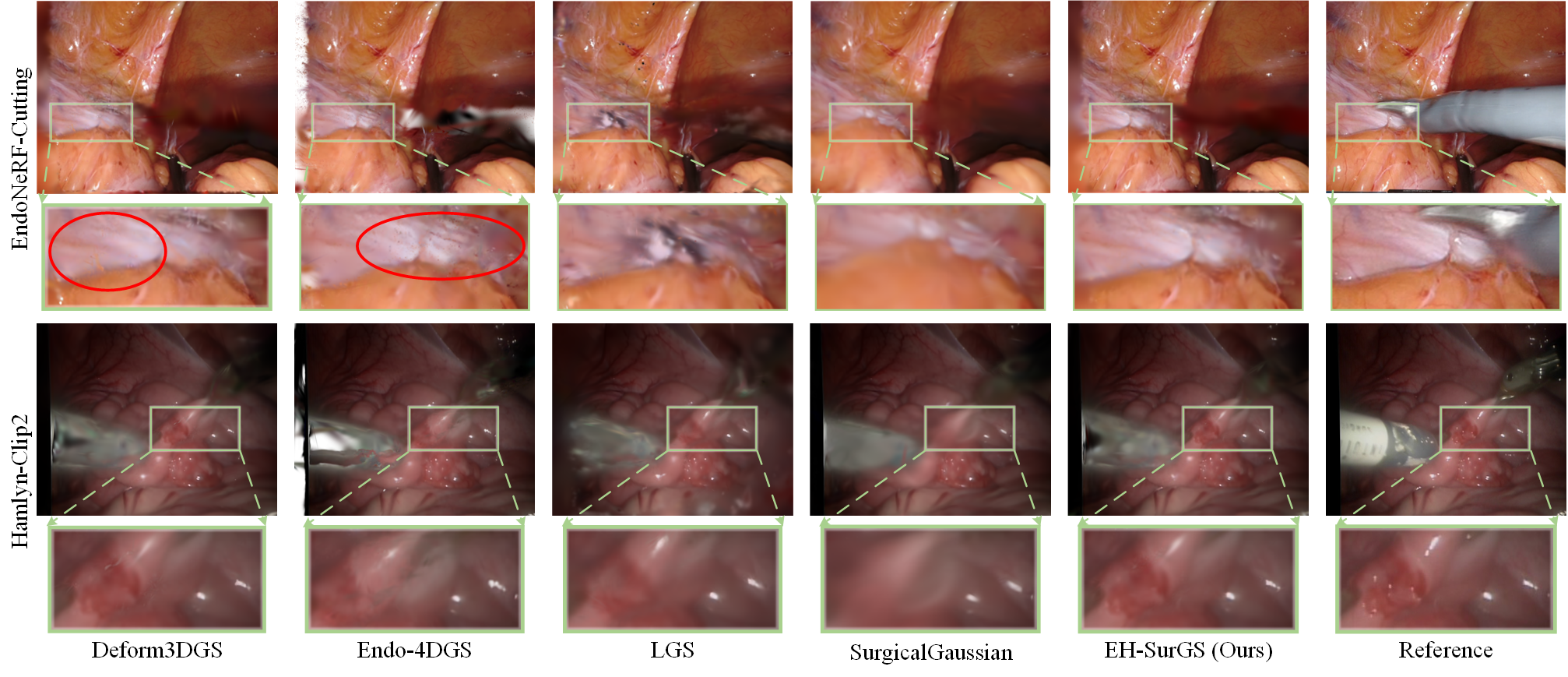}}
    \vspace{-4pt}
    \caption{Visualization of reconstruction results. The baselines show artifacts or blurriness in their reconstructions, while our approach achieves high-quality reconstruction performance.}
    \vspace{-10pt}
    \label{fig:result}
\end{figure*}
\subsection{Experimental Results}
We present quantitative results for the EndoNeRF dataset \cite{endonerf}, Hamlyn dataset, and StereoMIS dataset \cite{lyu2023learning} in Table~\ref{tab:qua}. EH-SurGS outperforms other state-of-the-art methods in both reconstruction quality and rendering speed, demonstrating significant superiority. Forplane \cite{yang2024efficient}, which is based on neural implicit representation, offers promising reconstruction quality but has a maximum rendering speed of only 17.15 fps. In contrast, the rendering speed of our EH-SurGS ranges from 349.00 fps to 400.00 fps, which is significantly higher than neural implicit representation methods. While the training time of Deform3DGS\cite{yang2024deform3dgs} is slightly better than ours, its performance is inferior to EH-SurGS across all datasets. Moreover, its rendering speed is significantly lower than ours. Endo-4DGS\cite{huang2024endo}, LGS\cite{liu2024lgs}, and SurgicalGaussian\cite{xie2024surgicalgaussian} also exhibit good performance, but EH-SurGS surpasses these methods in terms of reconstruction quality, training time, and rendering speed across all six datasets. This highlights the superior performance of our approach, emphasizing the effectiveness of deformation modeling with life cycle and adaptive motion hierarchy strategy. Fig.~\ref{fig:result} shows the qualitative results. Compared to baselines, EH-SurGS effectively reduces ghosts in the reconstructed scenes, resulting in finer and more accurate details. Notably, EH-SurGS achieves these improvements with shorter training time and faster rendering speed.
\subsection{Ablation Study}\label{sec:ablation}
We conduct an ablation study on the EndoNeRF-Cutting dataset. Table \ref{tab:ablation} compares the reconstruction quality and rendering speed of our method with various baseline methods.

\begin{figure}[t]
    \centerline{\includegraphics[width=1\linewidth]{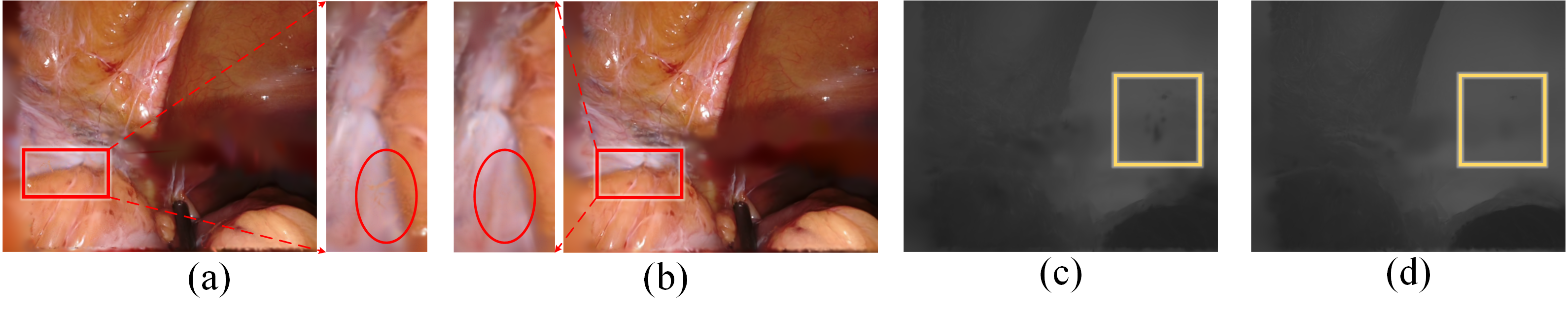}}
    \vspace{-8pt}
    \caption{{(a) The rendered RGB image using the w/o LC model; (b) The rendered RGB image using our proposed deformation model (Full); (c) The rendered depth map using the w/o LC-add model; (d) The rendered depth map using our proposed  deformation model (Full). Regions of interest are highlighted using red and yellow boxes.}}
    \vspace{-15pt}
    \label{fig:ablation}
\end{figure}

\textbf{Effectiveness of deformable modeling with life cycle.} As shown in Table~\ref{tab:ablation}, when using the existing deformable surgical scene modeling scheme without considering the life cycle of 3DGS (\textbf{w/o LC}), the reconstruction performance decreases. {\textbf{w/o LC-add} refers to modeling the life cycle via \(\alpha_t = \alpha_0 \times \sum_{j=1}^{B} \omega_j^{\alpha} b^{\alpha}(t)\), similar to \cite{li2024spacetime, zhao2024gaussianprediction}, instead of \eqref{equ:alpha}.} {While the performance under this approach improves over that of \textbf{w/o LC}, it still remains inferior to our proposed method.} {This is mainly because when \(\alpha_0\) is very small, the resulting \(\alpha_t\) is also very small. Therefore, this approach cannot effectively model newly appearing 3D Gaussians.} Fig.~\ref{fig:ablation} visualizes the RGB and depth maps rendered by different methods. Artifacts are evident in the results of \textbf{w/o LC} and \textbf{w/o LC-add}.
 \begin{table}[t]
 \vspace{-8pt}
\caption{The ablation study results for key components of EH-SurGS on the EndoNeRF-Cutting dataset\cite{endonerf}.}\label{tab:ablation}
    \centering
    \begin{tabularx}{\columnwidth}{c*{4}{>{\centering\arraybackslash}X}}
    \toprule
     & PSNR\(\uparrow\) & SSIM\(\uparrow\) & LPIPS\(\downarrow\) & FPS\(\uparrow\) \\
    \midrule
    Full&39.91&0.972&0.034&379.67 \\
    \midrule
    w/o LC&38.62&0.965&0.046&378.00 \\
    w/o LC-add&39.44&0.971&0.034&379.00 \\
    \midrule
    w/o AMHS&39.92&0.972&0.034&351.00 \\
    w/o UC1&39.91&0.972&0.034&357.33 \\
    w/o UC2&39.90&0.971&0.036&372.00 \\
    w/o RS&39.80&0.972&0.036&384.67 \\
    w/o MIU&39.85&0.971&0.036&376.67 \\
    \bottomrule
    \end{tabularx}
    \vspace{-15pt}
\end{table}

\textbf{Effectiveness of adaptive motion hierarchy strategy.} \textbf{w/o AMHS} removes the entire adaptive motion hierarchy strategy from our framework. The results show no decrease in reconstruction performance but a significant drop in rendering speed. This demonstrates the strategy's importance for improving rendering speed. We further explore key components of the adaptive motion hierarchy strategy. \textbf{w/o UC1} and \textbf{w/o UC2} represent the removal of update criteria 1 and 2 for the mask, respectively. As shown in Table~\ref{tab:ablation}, removing either criterion reduces rendering speed. This happens because a single criterion cannot accurately separate deformed and static regions. \textbf{w/o RS} means no further splitting is performed on conflict areas during mask updates. This categorizes conflict areas as static, increasing rendering speed but lowering reconstruction quality. \textbf{w/o MIU} maintains the default number of mask updates. This approach hinders training stability and results in lower reconstruction performance.
\section{Conclusion}
In this paper, we propose an efficient and high-fidelity reconstruction framework, EH-SurGS, for deformable surgical environments. Our design approach is twofold. First, to effectively model irreversible deformations, we introduce the concept of the life cycle for 3D Gaussians based on existing deformation modeling methods. This significantly improves reconstruction quality. Second, we propose an adaptive motion hierarchy strategy to distinguish between static and deformable areas in the scene. This strategy accelerates rendering speed. Evaluations on multiple in vivo datasets show that EH-SurGS outperforms other methods in both reconstruction performance and rendering speed. We believe that EH-SurGS has great potential for surgical applications.

{However, our method is currently designed for static endoscope settings, which limits its applicability to scenarios like colonoscopy. In future work, we plan to extend our framework to handle deformable scenes with moving endoscopes, making it more versatile for a broader range of medical endoscopic applications.}
\bibliographystyle{IEEEtran}
\bibliography{IEEEexample.bib}

\begin{thebibliography}{10}
\providecommand{\url}[1]{#1}
\csname url@samestyle\endcsname
\providecommand{\newblock}{\relax}
\providecommand{\bibinfo}[2]{#2}
\providecommand{\BIBentrySTDinterwordspacing}{\spaceskip=0pt\relax}
\providecommand{\BIBentryALTinterwordstretchfactor}{4}
\providecommand{\BIBentryALTinterwordspacing}{\spaceskip=\fontdimen2\font plus
\BIBentryALTinterwordstretchfactor\fontdimen3\font minus \fontdimen4\font\relax}
\providecommand{\BIBforeignlanguage}[2]{{%
\expandafter\ifx\csname l@#1\endcsname\relax
\typeout{** WARNING: IEEEtran.bst: No hyphenation pattern has been}%
\typeout{** loaded for the language `#1'. Using the pattern for}%
\typeout{** the default language instead.}%
\else
\language=\csname l@#1\endcsname
\fi
#2}}
\providecommand{\BIBdecl}{\relax}
\BIBdecl

\bibitem{tewari2020state}
A.~Tewari, O.~Fried, J.~Thies, V.~Sitzmann, S.~Lombardi, K.~Sunkavalli, R.~Martin-Brualla, T.~Simon, J.~Saragih, M.~Nie{\ss}ner \emph{et~al.}, ``State of the art on neural rendering,'' in \emph{Computer Graphics Forum}, vol.~39, no.~2.\hskip 1em plus 0.5em minus 0.4em\relax Wiley Online Library, 2020, pp. 701--727.

\bibitem{tewari2022advances}
A.~Tewari, J.~Thies, B.~Mildenhall, P.~Srinivasan, E.~Tretschk, W.~Yifan, C.~Lassner, V.~Sitzmann, R.~Martin-Brualla, S.~Lombardi \emph{et~al.}, ``Advances in neural rendering,'' in \emph{Computer Graphics Forum}, vol.~41, no.~2.\hskip 1em plus 0.5em minus 0.4em\relax Wiley Online Library, 2022, pp. 703--735.

\bibitem{mildenhall2021nerf}
B.~Mildenhall, P.~P. Srinivasan, M.~Tancik, J.~T. Barron, R.~Ramamoorthi, and R.~Ng, ``Nerf: Representing scenes as neural radiance fields for view synthesis,'' \emph{Communications of the ACM}, vol.~65, no.~1, pp. 99--106, 2021.

\bibitem{endonerf}
Y.~Wang, Y.~Long, S.~H. Fan, and Q.~Dou, ``Neural rendering for stereo 3d reconstruction of deformable tissues in robotic surgery,'' in \emph{International Conference on Medical Image Computing and Computer-Assisted Intervention}.\hskip 1em plus 0.5em minus 0.4em\relax Springer, 2022, pp. 431--441.

\bibitem{endosurf}
R.~Zha, X.~Cheng, H.~Li, M.~Harandi, and Z.~Ge, ``Endosurf: Neural surface reconstruction of deformable tissues with stereo endoscope videos,'' in \emph{International Conference on Medical Image Computing and Computer-Assisted Intervention}.\hskip 1em plus 0.5em minus 0.4em\relax Springer, 2023, pp. 13--23.

\bibitem{yang2023neural}
C.~Yang, K.~Wang, Y.~Wang, X.~Yang, and W.~Shen, ``Neural lerplane representations for fast 4d reconstruction of deformable tissues,'' \emph{arXiv preprint arXiv:2305.19906}, 2023.

\bibitem{batlle2023lightneus}
V.~M. Batlle, J.~M. Montiel, P.~Fua, and J.~D. Tard{\'o}s, ``Lightneus: Neural surface reconstruction in endoscopy using illumination decline,'' in \emph{International Conference on Medical Image Computing and Computer-Assisted Intervention}.\hskip 1em plus 0.5em minus 0.4em\relax Springer, 2023, pp. 502--512.

\bibitem{10542414}
J.~Shan, Y.~Li, T.~Xie, and H.~Wang, ``Enerf-slam:a dense endoscopic slam with neural implicit representation,'' \emph{IEEE Transactions on Medical Robotics and Bionics}, vol.~6, no.~3, pp. 1030--1041, 2024.

\bibitem{yang2024deform3dgs}
S.~Yang, Q.~Li, D.~Shen, B.~Gong, Q.~Dou, and Y.~Jin, ``Deform3dgs: Flexible deformation for fast surgical scene reconstruction with gaussian splatting,'' \emph{arXiv preprint arXiv:2405.17835}, 2024.

\bibitem{huang2024endo}
Y.~Huang, B.~Cui, L.~Bai, Z.~Guo, M.~Xu, and H.~Ren, ``Endo-4dgs: Distilling depth ranking for endoscopic monocular scene reconstruction with 4d gaussian splatting,'' \emph{arXiv preprint arXiv:2401.16416}, 2024.

\bibitem{li2024endosparse}
C.~Li, B.~Y. Feng, Y.~Liu, H.~Liu, C.~Wang, W.~Yu, and Y.~Yuan, ``Endosparse: Real-time sparse view synthesis of endoscopic scenes using gaussian splatting,'' \emph{arXiv preprint arXiv:2407.01029}, 2024.

\bibitem{liu2024lgs}
H.~Liu, Y.~Liu, C.~Li, W.~Li, and Y.~Yuan, ``Lgs: A light-weight 4d gaussian splatting for efficient surgical scene reconstruction,'' \emph{arXiv preprint arXiv:2406.16073}, 2024.

\bibitem{xie2024surgicalgaussian}
W.~Xie, J.~Yao, X.~Cao, Q.~Lin, Z.~Tang, X.~Dong, and X.~Guo, ``Surgicalgaussian: Deformable 3d gaussians for high-fidelity surgical scene reconstruction,'' \emph{arXiv preprint arXiv:2407.05023}, 2024.

\bibitem{kerbl20233d}
B.~Kerbl, G.~Kopanas, T.~Leimk{\"u}hler, and G.~Drettakis, ``3d gaussian splatting for real-time radiance field rendering.'' \emph{ACM Trans. Graph.}, vol.~42, no.~4, pp. 139--1, 2023.

\bibitem{yang2024efficient}
C.~Yang, K.~Wang, Y.~Wang, Q.~Dou, X.~Yang, and W.~Shen, ``Efficient deformable tissue reconstruction via orthogonal neural plane,'' \emph{IEEE Transactions on Medical Imaging}, 2024.

\bibitem{zwicker2002ewa}
M.~Zwicker, H.~Pfister, J.~Van~Baar, and M.~Gross, ``Ewa splatting,'' \emph{IEEE Transactions on Visualization and Computer Graphics}, vol.~8, no.~3, pp. 223--238, 2002.

\bibitem{li2024spacetime}
Z.~Li, Z.~Chen, Z.~Li, and Y.~Xu, ``Spacetime gaussian feature splatting for real-time dynamic view synthesis,'' in \emph{Proceedings of the IEEE/CVF Conference on Computer Vision and Pattern Recognition}, 2024, pp. 8508--8520.

\bibitem{zhao2024gaussianprediction}
B.~Zhao, Y.~Li, Z.~Sun, L.~Zeng, Y.~Shen, R.~Ma, Y.~Zhang, H.~Bao, and Z.~Cui, ``Gaussianprediction: Dynamic 3d gaussian prediction for motion extrapolation and free view synthesis,'' in \emph{ACM SIGGRAPH 2024 Conference Papers}, 2024, pp. 1--12.

\bibitem{wang2023sparsenerf}
G.~Wang, Z.~Chen, C.~C. Loy, and Z.~Liu, ``Sparsenerf: Distilling depth ranking for few-shot novel view synthesis,'' in \emph{Proceedings of the IEEE/CVF International Conference on Computer Vision}, 2023, pp. 9065--9076.

\bibitem{hayoz2023learning}
M.~Hayoz, C.~Hahne, M.~Gallardo, D.~Candinas, T.~Kurmann, M.~Allan, and R.~Sznitman, ``Learning how to robustly estimate camera pose in endoscopic videos,'' \emph{International journal of computer assisted radiology and surgery}, vol.~18, no.~7, pp. 1185--1192, 2023.

\bibitem{recasens2021endo}
D.~Recasens, J.~Lamarca, J.~M. F{\'a}cil, J.~Montiel, and J.~Civera, ``Endo-depth-and-motion: Reconstruction and tracking in endoscopic videos using depth networks and photometric constraints,'' \emph{IEEE Robotics and Automation Letters}, vol.~6, no.~4, pp. 7225--7232, 2021.

\bibitem{xiong2024efficientsam}
Y.~Xiong, B.~Varadarajan, L.~Wu, X.~Xiang, F.~Xiao, C.~Zhu, X.~Dai, D.~Wang, F.~Sun, F.~Iandola \emph{et~al.}, ``Efficientsam: Leveraged masked image pretraining for efficient segment anything,'' in \emph{Proceedings of the IEEE/CVF Conference on Computer Vision and Pattern Recognition}, 2024, pp. 16\,111--16\,121.

\bibitem{ssim}
Z.~Wang, A.~C. Bovik, H.~R. Sheikh, and E.~P. Simoncelli, ``Image quality assessment: from error visibility to structural similarity,'' \emph{IEEE transactions on image processing}, vol.~13, no.~4, pp. 600--612, 2004.

\bibitem{lpips}
R.~Zhang, P.~Isola, A.~A. Efros, E.~Shechtman, and O.~Wang, ``The unreasonable effectiveness of deep features as a perceptual metric,'' in \emph{Proceedings of the IEEE conference on computer vision and pattern recognition}, 2018, pp. 586--595.

\bibitem{lyu2023learning}
X.~Lyu, P.~Dai, Z.~Li, D.~Yan, Y.~Lin, Y.~Peng, and X.~Qi, ``Learning a room with the occ-sdf hybrid: Signed distance function mingled with occupancy aids scene representation,'' \emph{arXiv preprint arXiv:2303.09152}, 2023.

\end{thebibliography}
\end{document}